# Which2comm: An Efficient Collaborative Perception Framework for 3D Object Detection

Duanrui Yu[1*]  Jing You[1*]  Xin Pei[1†]  Anqi Qu[1]  Dingyu Wang[1]  Shaocheng Jia[2]
[1]Department of Automation, BNRIST, Tsinghua University
[2]Department of Civil Engineering, The University of Hong Kong

[*]Equal Contribution  [†]Corresponding Author

## Abstract

*Collaborative perception allows real-time inter-agent information exchange and thus offers invaluable opportunities to enhance the perception capabilities of individual agents. However, limited communication bandwidth in practical scenarios restricts the inter-agent data transmission volume, consequently resulting in performance declines in collaborative perception systems. This implies a trade-off between perception performance and communication cost. To address this issue, we propose **Which2comm**, a novel multi-agent 3D object detection framework leveraging object-level sparse features. By integrating semantic information of objects into 3D object detection boxes, we introduce semantic detection boxes (SemDBs). Innovatively transmitting these information-rich object-level sparse features among agents not only significantly reduces the demanding communication volume, but also improves 3D object detection performance. Specifically, a fully sparse network is constructed to extract SemDBs from individual agents; a temporal fusion approach with a relative temporal encoding mechanism is utilized to obtain the comprehensive spatiotemporal features. Extensive experiments on the V2XSet and OPV2V datasets demonstrate that Which2comm consistently outperforms other state-of-the-art methods on both perception performance and communication cost, exhibiting better robustness to real-world latency. These results present that for multi-agent collaborative 3D object detection, transmitting only object-level sparse features is sufficient to achieve high-precision and robust performance.*

## 1. Introduction

Multi-agent collaborative perception allows information exchange among multiple agents through inter-terminal communication, effectively mitigating the occlusion issue of individual agents [1-5]. Thus, it has been widely applied in various fields [6-8] to enhance the perception performance, such as 3D object detection [9-11]. Particularly, Vehicle-to-Everything (V2X) technologies enables the information sharing between vehicles and roadside infrastructures. This facilitates connected and automated vehicles (CAVs) to gain long-distance perception and significantly improve traffic safety [4, 12, 13].

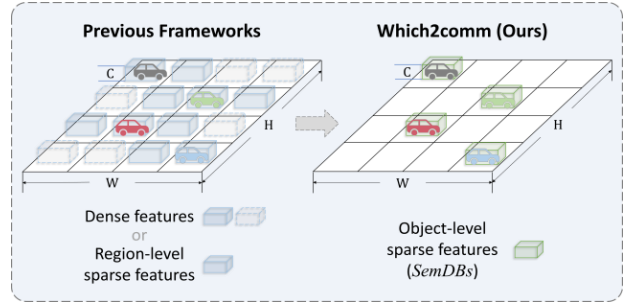

Figure 1. Which2comm innovatively transmits object-level sparse features solely leveraging SemDBs, which significantly reduces the demanding communication volume. In the figure, H, W and C represent the height, weight of detection boxes and feature dimensionality, respectively.

However, in real-world scenarios, the performance of collaborative perception systems is constrained by additional factors apart from those in single-agent perception [10, 12]. For instance, as collaborative perception is built upon inter-agent information exchange, communication cost must be considered in collaborative perception algorithms. The communication bandwidth between agents is often limited in practical scenarios. Taking current vehicular communication standards as an example, the maximum available V2X communication bandwidth is 27 MB/s [14, 15], which cannot satisfy the data transmission requirements of most V2X collaborative perception algorithms. Given the limited bandwidth, it is challenging to transmit complete information among agents. Thus, the collaborative perception performance is significantly degraded with low communication volumes [10]. This presents an inevitable trade-off between perception performance and communication costs [4, 5, 13, 16] for collaborative perception methods.

To reduce communication cost, one of the effective approaches is the late fusion, which only transmits the output detection box reported by different agents [13]. However, due to the lack of semantic information, ambiguity arises during detection box deduplication from multiple agents under latency or localization errors, leading to inferior robustness. Therefore, current collaborative perception methods primarily focus on achieving the balance between bandwidth and performance through the intermediate fusion approach [1, 4, 5, 9, 16, 22, 39], which is charactered by preserving intermediate network features with semantic information for transmission. Such an approach reduces transmission cost via feature compression or sparsification, e.g., V2X-ViT [16] (it transmits compressed dense features) and Where2comm [4] (it transmits region-level sparse features). Nevertheless, excessive compression of features frequently leads to drastic detection performance drop. Moreover, the feature sparsification method still remains at the regional level [4, 9], indicating significant potential for further exploration in this domain.

In this paper, we focus on leveraging the sparsity of observation data and constructing object-level sparse features for transmission to reduce communication costs. Considering the advantages of late fusion (in communication efficiency) and intermediate fusion (in the precision and robustness), we propose to combine their strengths by incorporating semantic information of objects into 3D detection boxes, forming *semantic detection boxes (SemDBs)*—an object-level sparse feature for transmission and fusion between agents and significantly reducing feature transmission cost (illustrated in Fig. 1).

Although the transmission volume can be effectively decreased, the data transmission time among agents is unavoidable, imposing negative impacts on perception performance. Therefore, taking the inevitable latency into consideration, we use the temporal information to enhance the detection performance and robustness. Prior works such as SCOPE[39] and CoBEVFlow[7] have achieved temporal fusion through different feature representation and integration architectures. To specifically accommodate sparse features, we adopt a relative temporal encoding (RTE) mechanism to explicitly distinguish features from different timestamps and effectively integrates the time-aware information. The temporal fusion approach enables the integration of dynamic information and effectively enhances the information density and reliability of SemDBs.

Drawing from the aforementioned considerations, we propose *Which2comm*, a novel multi-agent 3D object detection framework leveraging sparse SemDB. Figure 2 illustrates the structure of *Which2comm*, which includes the following modules:

1) *Extractor*. The sparse feature extraction module that mainly consists of sparse convolution structures and extracts SemDBs from the point clouds;

2) *Fuser*. The sparse feature fusion module that incorporates RTE into transmitted features and merges temporal features of multiple agents;

and 3) *Decoder*. The sparse feature decoding module that decodes explicit 3D object detection boxes based on the fused sparse features.

In addition, we further propose a shuffle-ego data augmentation strategy to take full advantage of the diversity of training data, thereby improving training effectiveness.

Comprehensive experiments conducted on the V2XSet [16] and OPV2V [23] datasets demonstrate that the proposed *Which2comm* outperforms the state-of-the-art (SOTA) methods in terms of the trade-off between perception performance and communication cost. *Which2comm* also exhibits better robustness to latency errors.

The main contributions of this paper can be summarized as follows:

• We propose *Which2comm*, a novel multi-agent 3D object detection framework that significantly reduces the inter-agent communication cost and improve perception performance.

• The proposed *Which2comm* achieves state-of-the-art detection performance and excellent robustness on latency. This demonstrates that for multi-agent collaborative 3D object detection, transmitting only object-level sparse features is sufficient to achieve high-precision and robust performance.

## 2. Related works

### 2.1. Collaborative perception

Multi-agent collaborative perception addresses the issue of limited view of individual agent, through information exchange of multiple agents with the support of intelligent connected technology. Various large-scale open datasets have emerged, promoting the development of perception frameworks [16, 23, 38]. According to the collaboration stage, frameworks for multi-agent collaborative 3D object detection can be categorized into three categories: early fusion that transmits raw sensor data between agents [24]; late fusion that transmits agents' detection outputs; and intermediate fusion that transmits intermediate features of neural networks [1, 4, 5, 9, 16-20, 39]. Early fusion performs significant advantage in detection precision but incurs higher data transmission costs; late fusion offers advantages in terms of data transmission but exhibits poorer robustness against latency and localization errors. Intermediate fusion has become a focal point in research due to its precision and robustness. Studies have emerged focusing on intermediate fusion methods, with a particular emphasis on addressing the issues of the bandwidth-performance trade-off [4, 5, 9, 19], inter-terminal latency [7], localization error [25], and so on. To achieve the bandwidth-performance trade-off, When2com [19] proposes a handshake mechanism to select communication partners; V2X-ViT [16] encode inter-terminal latency and utilize attention mechanisms to achieve structured intermediate fusion among multiple agents; SCOPE [39] effectively fuse the historical information by importance-aware adaptive fusion. The above methods transmit dense features, whereas Where2comm [4] uses a spatial confidence map for transmission, achieving region-level feature sparsification. CodeFilling [5] significantly reduces communication volume by maintaining a codebook among agents, and transmitting only integer indices of features, which markedly improves the algorithm's performance under limited communication bandwidth.

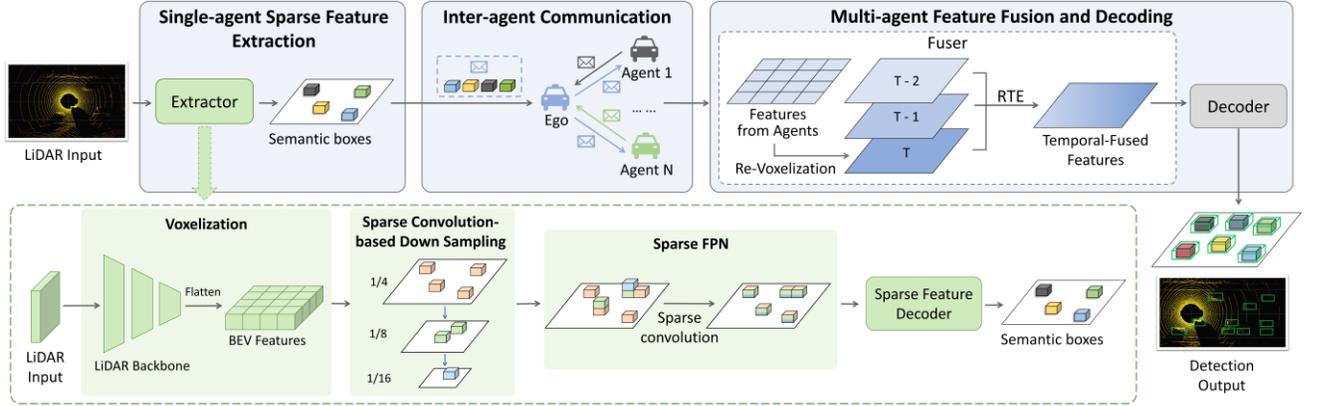

Figure 2. Overview of **Which2comm**, a multi-agent 3D object detection framework based on object-level sparse features. LiDAR point cloud is input into *Extractor*, obtaining the sparse SemDBs. Multi-agent sparse features are encoded with RTE, and further processed through *Fuser* and *Decoder*.

However, the above methods still lack effective utilization of the sparsity of observation, leaving room for improvement in data transmission volume. Approaches that emphasize feature compression often incur precision losses; methods transmitting region-level sparse features still retain redundant information; and maintaining a shared codebook among vehicles is challenging to apply to dynamic real-world scenarios. In this paper, we leverage the sparsity of LiDAR point cloud observation with a fully sparse network, obtaining object-level sparse features for transmission and feature fusion, and enhance performance by utilizing temporal information.

## 2.2. Sparse convolution

In many practical applications, input data frequently exhibits sparsity, which means that some elements are zero or inactive. Traditional fully connected convolution does not account for this sparsity, resulting in a significant computational waste on zero values. To address the inefficiency of traditional convolution when dealing with sparse data, researchers propose sparse convolution structures [26-30], These structures operate solely on non-zero or activated input elements, effectively reducing computational load. With the refinement of sparse convolution theory, various sparse convolution variants have been proposed, such as submanifold sparse convolution [28], focal sparse convolution [29] and convolution sparse encoding [30]. These innovations enhance the efficiency of sparse convolutions and tailor them to diverse application scenarios.

In the field of 3D object detection, sparse convolution is used to process LiDAR point cloud data. By directly detecting objects from sparse 3D data using fully sparse structures, FSD [32] and VoxelNeXt [33] perform improved computational efficiency and inference speed.

In this paper, we employ sparse convolution, submanifold sparse convolution and submanifold sparse max-pooling structures to build the fully sparse feature extraction module, *Extractor*, in order to preserve feature sparsity during computation and extract sparse SemDBs.

## 3. Which2comm

This section presents *Which2comm*, a multi-agent 3D object detection framework based on SemDB, which can effectively use the sparsity of point clouds and reduce transmission cost while enhancing performance. Section 3.1 first introduces the overall architecture. Details of the key modules are presented in Sections 3.2-3.3. Section 3.4 describes the proposed shuffle-ego training strategy. Section 3.5 offers the loss functions used in this paper.

### 3.1. Overall architecture

*Which2comm* comprises two main stages: single-agent sparse feature extraction and multi-agent temporal feature fusion and decoding. During the extraction stage, a fully sparse feature extraction module, *Extractor*, is constructed. Each individual agent obtains SemDBs. These sparse features are subsequently transmitted among terminals. During the multi-agent perception stage, SemDBs are enriched with temporal encoding, and further fed to *Fuser* and *Decoder* to obtain 3D object detection boxes.

**Sparse feature extraction.** Single-agent sparse feature extraction is applied to the input unstructured point cloud data, $X_i \in \mathbb{R}^{n_p \times 4}$, to obtain sparse features, where $n_p$ is the number of points and 4 represents the number of channels (i.e., 3D coordinate *(x, y, z)* and intensity). Let *N* be the total number of agents. For the *i* th agent, the *Extractor* module can be expressed as

$$F_i = \Phi_{extr}(X_i), \tag{1}$$

where $F_i \in \mathbb{R}^{m_i \times c}$ is the extracted sparse feature map, namely SemDBs; $\Phi_{extr}(\cdot)$ is an operator representing the extraction process; and $m_i$ and *c*, respectively, represent the number of detected objects and the feature dimension.

**Sparse feature transmission.** Agents interact with each other by exchanging information through data transmission. During this progress, the transmission cost from collaborator $j$ to ego agent $i$ for a single frame can be computed as:
$$C_{j \to i}^{tran} = m_j \times \left(\dim(F_j) + \dim(I_j^{spac})\right), \quad (2)$$
where $\dim(F_j)$ and $\dim(I_j^{spac})$ represent the dimensions of transmitted features and the corresponding spatial indices, respectively. The set of features detected and received by ego agent $i$ is denoted as $\widetilde{F}_i$, thus the transmission process can be represented as:
$$\widetilde{F}_i = \cup_{j, j \neq i}(F_{j \to i}), \quad (3)$$

**Sparse feature fusion.** *Fuser* module is used to incorporate temporal encoding into sparse features from multiple agents and further combine features from the past $p$ time steps. For ego agent $i$, denote $f_{tmp\_enc}(\cdot)$ and $\Phi_{fus}(\cdot)$ as the temporal encoding mechanism and the fusion operator, respectively. The *Fuser* module can be formulated as follows:
$$T_{i,j}^{t_k} = f_{tmp\_enc}\left(\widetilde{F}_{i,j}^{t_k}\right), \quad (4)$$
$$H_i = \Phi_{fus}\left(\left\{T_{i,j}^{t_k}\right\}_{k=c,c-1,\ldots,c-p; j=0,1,\ldots,N}\right), \quad (5)$$
where $t_k \in \{t_c, t_{c-1}, \ldots, t_{c-p}\}$ is the temporal index representing the $k$-th timestamp earlier than current time $t_c$; $\widetilde{F}_{i,j}^{t_l} \in \mathbb{R}^{m_j \times c}$ is the features transmitted from agents $j$ to $i$ at $t_i$; $T_{i,j}^{t_k} \in \mathbb{R}^{m_j \times c}$ is the corresponding temporally encoded features; and $H_i \in \mathbb{R}^{\widetilde{m}_i \times c}$ represents the fused sparse features of agent $i$ with the final object count of $\widetilde{m}_i$.

**Sparse feature decoding.** *Decoder* is used to process the fused features, $H_i$, and convert the implicit sparse semantic features into the explicit 3D object detection boxes. This can be expressed as:
$$B_i = \Phi_{dec}(H_i), \quad (6)$$
where $\Phi_{dec}(\cdot)$ represents the *Decoder* function; and $B_i \in \mathbb{R}^{\widetilde{m}_i \times 7}$ is the output 3D detection boxes of ego agent; $\widetilde{m}_i$ denotes the number of 3D detection boxes; and 7 represents the number of attributes of each 3D detection box (i.e., the 3D coordinate of box center, $(x, y, z)$, and the height, width, and length of the box, $(h, w, l)$, and the yaw rotation angle).

### 3.2. Single-agent sparse feature extraction

*Extractor* preserves sparsity throughout the feature propagation process by employing a fully sparse structure. In this module, the input point clouds $X_i$ are initially converted into voxel-level features through voxelization structures, then down-sampled through consecutive sparse convolution modules to capture features at various scales. Subsequently, a sparse feature pyramid network (sparse FPN) is employed for multi-level feature fusion, and SemDBs are generated through a carefully-designed sparse feature decoding module.

**Voxel feature encoding.** 2D voxelization is applied to transform the input point clouds $X_i$ into voxel-level features, denoted as $V_i \in \mathbb{R}^{n_v \times c}$, where $n_v$ is the number of voxels. Additionally, voxel indices $I_i \in \mathbb{R}^{n_v \times 4}$ are provided for subsequent convolutions.

**Multi-scale sparse feature extraction.** Multi-scale sparse feature extraction is used to extract object information from voxel features and capturing features at various scales while preserving sparsity. This extraction process involves stacked sparse convolutional building blocks containing a sparse convolutional layer and multiple submanifold sparse convolutional structures. To alleviate the degradation phenomenon, residual connection structures akin to ResNet [34] are integrated into sparse convolutional layers. As features propagate in the network, the receptive field of the features continuously increases, deepening the level of abstraction. Ultimately, features down-sampled by factors of 1×, 2×, 4×, 8×, and 16× are obtained, denoted as $F_i^{ds\_1\times}, F_i^{ds\_2\times}, F_i^{ds\_4\times}, F_i^{ds\_8\times}$ and $F_i^{ds\_16\times}$, respectively.

**Sparse feature pyramid network.** The multi-scale features $F_i^{ds\_4\times}, F_i^{ds\_8\times}$, and $F_i^{ds\_16\times}$ are fused through sparse FPN to obtain 4× down-sampled features. Sparse FPN is transitioned from the conventional feature pyramid network (FPN) by replacing the operators handling dense features with the corresponding sparse versions, as illustrated in Fig. 3(a). To further process the fused sparse features, spatial index up-sample is employed, which solely up-samples the spatial indices of higher-level features without altering the values. The up-sampled sparse features with identical positional indices are directly summed to merge features from different levels, referred to as sparse addition. The sparse FPN layer outputs fused features that match the dimensions of the 4× down-sampled features, denoted as $F_i^{4\times}$.

**Sparse feature decoding.** The decoding structure comprises a confidence head and a box center head, enabling three critical processes: **box center adjustment, box selection**, and **feature re-weighting**. The decoding heads are constructed with sparse convolutional layers, predicting the classification confidences $\{Y_{i,k}^{conf}\}_{i=1,2,\ldots,N; k=1,2,\ldots,m_i}$ and the relative positional deviations between sparse features and the actual detection box center $\{Y_{i,k}^{dev}\}_{i=1,2,\ldots,N; k=1,2,\ldots,m_i}$, respectively.

As LiDAR point clouds mainly capture the outer surfaces of objects, sparse feature positions often deviate from the actual center of 3D object detection box. To mitigate the negative impact on feature fusion across agents, a box center head is introduced to predict the positional deviation between sparse features and box centers. The adjusted spatial index of sparse feature is calculated as $\hat{I}_i^k = I_i^k + Y_{i,k}^{dev}$, hence the original index plus the relative deviation.

Apart from the required object features, sparse features obtained through sparse FPN also contains background features and redundant features. Features with confidence $Y_{i,k}^{conf}$ below the threshold are considered background and excluded from the candidate set to maintain the sparsity. Meanwhile, a submanifold sparse max-pooling [28] layer is employed to remove redundant features, which serves the function instead of Non-Maximum Suppression (NMS), yet suitable for sparse features.

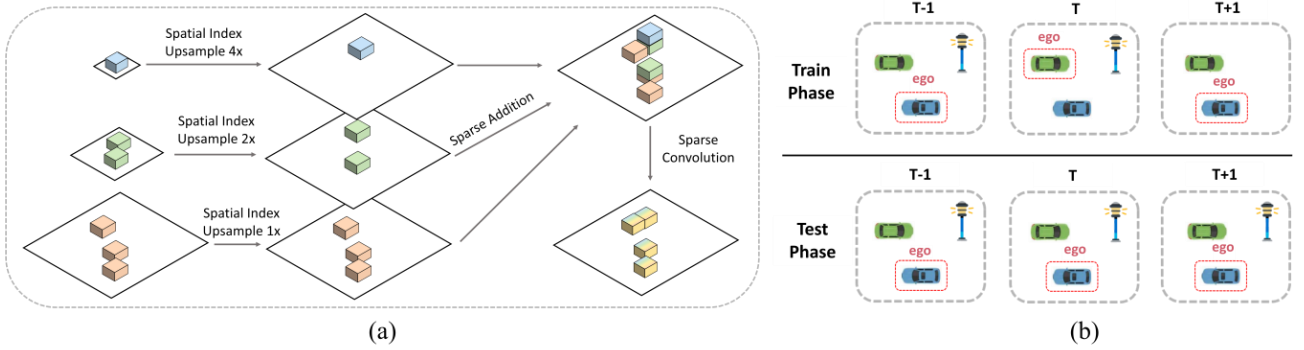

Figure 3. (a) Sparse feature pyramid network. Differently colored blocks denote various levels of sparse features, with differently sized background squares indicating the down-sampled feature space. (b) Shuffle-ego data augment strategy.

Moreover, features from diverse agents can be influenced by noise and uncertainties. Feature re-weighting is used to leverage this characteristic. The sparse features are embedded with uncertainties, enabling a more precise capture and utilization of the reliability of distinct features. Specifically, sparse features are multiplied by the predicted confidence $Y_{i,k}^{conf}$ to suppress features with higher uncertainties. The ablation experiment results in Section 4.4 demonstrate that the feature re-weighting is an effective strategy to enhance model performance.

Through our aforementioned extraction pipeline, multi-scale features of objects are extracted, fused, and critically filtered and refined, resulting in highly condensed and information-rich object-level sparse representations, namely SemDBs. Each SemDB provides sufficient semantic information for a single detection box, offering significant advantages over previous dense and region-level sparse features. This semantic richness, enabled by our meticulously designed *Extractor*, leads to superior detection performance while maintaining communication efficiency. Fig. 5(a) visualizes these SemDBs, with further analysis provided in Section 4.3.

### 3.3. Multi-agent feature fusion and decoding

Agents exchange the obtained SemDBs during the inter-agent communication phase. Subsequently, the sparse feature fusion module, *Fuser*, is utilized to obtain the fused sparse features $H_i$. Afterwards, the explicit 3D object detection boxes $B_i$ are obtained through the decoding module of sparse features, referred to as *Decoder*.

**Relative temporal encoding mechanism.** To differentiate features from different timestamps, RTE mechanism is utilized. Inspired by positional encoding mechanism in Transformer [35], the RTE is represented as:

$$F_{RTE}(\Delta t, 2k) = \sin\left(\frac{\Delta t}{10000^{2k/D}}\right), \quad (7)$$

$$F_{RTE}(\Delta t, 2k+1) = \cos\left(\frac{\Delta t}{10000^{2k/D}}\right), \quad (8)$$

where $\Delta t$ represents the difference between historical and current features' timestamp; $D$ and $k$ respectively denote the dimension of features and the index of channel within the feature. The timestamp-aware vector $F_{RTE}(\Delta t, k)$ is added to features directly, obtaining temporally encoded vector, $T_{i,j}^{t_k}$.

**Re-voxelization.** In the *Fuser* module, features collected from different agents and timestamps are fused through sparse addition. However, due to the lower voxel resolution after down-sampling, sparse features from different time steps frequently map to the same voxel. To address this, we employ a re-voxelization method that enhances voxel resolution and reduces feature overlapping.

**Feature fusion.** The re-voxelized features encoded with RTE are fed to a sparse addition module to fuse features with the same spatial index across different time steps. Subsequently, further fusion is conducted through down-sampling and sparse FPN, reusing the structures within *Extractor*. The fused features $H_i$ are obtained.

**Feature decoding.** *Decoder* consists of four feature decoding heads: confidence head, box center head, box dimension head, and box rotation head, which share nearly identical structures. Confidence head outputs the detection confidence; box center head outputs the deviations in $x$, $y$, and $z$ directions; box dimension head outputs the length, width, and height of the 3D object detection box; and box rotation head outputs the rotation angle $yaw$ around the $z$-axis. Combining and transforming this information yields the output 3D object detection boxes $B_i$.

### 3.4. Shuffle-ego data augment strategy

Multi-agent datasets contain data from multiple agents within the same scene. During the training and inference processes of multi-agent object detection models, it is necessary to specify one agent as the reference center, known as the ego agent. When evaluating the effectiveness of models, a fixed ego agent is chosen to establish a unified benchmark for comparison. However, in training, mainstream methods [1, 4, 16] have also adopted this setting. This approach does not fully exploit the potential of multi-terminal datasets and results in lower utilization rates.

Table 1. Quantitative comparison results of detection performance and communication cost on V2XSet and OPV2V.

| Dataset | V2XSet | | | OPV2V | | |
|---|---|---|---|---|---|---|
| Method/Metric | AP@0.5↑ | AP@0.7↑ | AB↓ | AP@0.5↑ | AP@0.7↑ | AB↓ |
| No Fusion | 0.606 | 0.402 | 0 | 0.679 | 0.602 | 0 |
| Early Fusion | 0.881 | 0.793 | 27.480 | 0.908 | 0.854 | 28.001 |
| Late Fusion | 0.776 | 0.649 | 6.477 | 0.885 | 0.824 | 6.573 |
| V2VNet [1] | 0.866 | 0.690 | 23.883 | 0.917 | 0.822 | 23.779 |
| V2X-ViT [16] | 0.882 | 0.713 | 18.735 | 0.866 | 0.769 | 18.779 |
| Where2comm [4] | 0.833 | 0.624 | 22.004 | 0.877 | 0.775 | 22.899 |
| CoAlign [22] | 0.881 | 0.774 | 23.883 | 0.903 | 0.833 | 23.779 |
| SCOPE [39] | 0.877 | 0.764 | 21.251 | 0.898 | 0.811 | 21.033 |
| CodeFilling [5] (max-vol) | 0.910 | 0.825 | 15.802 | 0.919 | 0.842 | 15.796 |
| CodeFilling [5] (matched-vol) | 0.909 | 0.821 | 14.562 | 0.916 | 0.841 | 13.924 |
| **Which2comm (Ours)** | **0.929** | **0.833** | **14.487** | **0.925** | **0.873** | **13.834** |

The shuffle-ego data augment strategy is illustrated in Fig. 3(b). The shown scene contains 3 agents. During the testing phase, the blue vehicle is consistently chosen as the ego agent. However, during training, the selection of the ego agent is random, switching between the blue and green vehicles at different time points. The inclusion of each agent's data in the training process is also randomized: at times, roadside devices are included in the training data, while at other times, they are excluded.

### 3.5. Loss functions

The decoding heads in both *Extractor* and *Decoder* can be classified into two categories: classification heads and regression heads. We use Focal Loss [36] for the classification task, represented as $L_{cla}$ and $L1$ loss for the regression task, represented as $L_{reg}$. The overall loss function is the weighted sum of individual loss functions of each decoding head, namely $L = L_{cla} + \alpha L_{reg}$, where $\alpha$ is set to 1 in the experiments. The final loss is computed as the direct sum of losses from *Extractor* and *Decoder* heads.

## 4. Experiments

### 4.1. Datasets and experimental settings

**Datasets**. V2XSet [16] is a large-scale simulation dataset specifically designed for V2X perception, well-suited for exploring typical challenges including communication latency. The total 11,447 frames of perception scenarios contain CAVs as well as roadside units and is divided into train/validation/test splits of 6,694/1,920/2,833 frames. OPV2V [23] is a cooperative Vehicle-to-Vehicle perception dataset containing over 70 scenes, which provides a variety of collaborative perception tasks in different traffic scenarios. The train/validation/test splits of OPV2V are 6764/1981/2709 frames. To ensure an objective comparison of perception methods, we utilized a singular LiDAR point cloud modality as the input for all the methods.

**Metrics**. The evaluation focuses on both perception performance and communication efficiency of baselines and SOTA collaborative 3D object detection frameworks. The perception performance is evaluated with average precision (AP) at IoU 0.5 and 0.7 (represented as AP@0.5 and AP@0.7). The communication efficiency is evaluated with average byte (AB), computing the bytes of data transmitted from collaborators to ego vehicle, averaged across all frames. To align with the commonly used metrics, we calculate the megabyte-based volume on $log_2$ scale. For example, 1 megabyte (MB) of transmission volume corresponds to an AB of 20. According to Eq. (2), volume of *Which2comm* in a single frame can be calculated as $\log_2\left(\Sigma_{j, j \neq i}(C + 2) \times m_j \times 32/8\right)$, where $C$ and $m_j$ denotes the number of channels and SemDBs agent $j$ detected, respectively, 2 is the dimension of spatial indices, and 32/8 represents the float32 data calculated in bytes.

**Experimental settings**. The parameter settings follow [23]. The LiDAR range for CAVs is $x \in [-140.8m, +140.8m]$, $y \in [-40m, +40m]$. The tested models share LiDAR backbone of PointPillars [37]. In terms of *Which2comm*, we use the model that incorporates two historical frames (*p*=2).

### 4.2. Quantitative results

**Benchmark comparison**. Tab. 1 presents the overall performance of collaborative 3D object detection methods on V2XSet and OPV2V. We conduct the comparison on baseline models including no fusion, early fusion, and late fusion, as well as SOTA intermediate fusion models including V2VNet [1], V2X-ViT [16], Where2comm [4], CoAlign [22], SCOPE [39] and CodeFilling [5]. As for CodeFilling, we present the result under its maximum volume (CodeFilling (max-vol)) as well as the result under the volume consistent with that of *Which2comm* (CodeFilling (matched-vol)), in order to compare both detection performance and transmission cost. The evaluation results indicate that the proposed *Which2comm* significantly outperforms previous methods in **both detection performance and communication cost**. In terms of detection performance, *Which2comm* achieves SOTA performance on both datasets, outperforming SCOPE [39], which incorporates historical information by adaptive fusion, as well as the previous SOTA, CodeFilling [5].

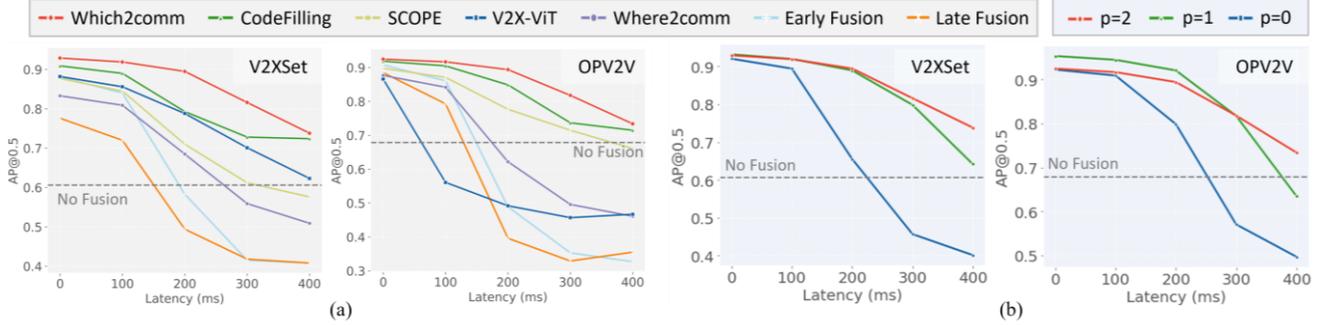

Figure 4. (a) Evaluation of latency robustness on V2XSet and OPV2V. (b) Ablation of the temporal fusion mechanism.

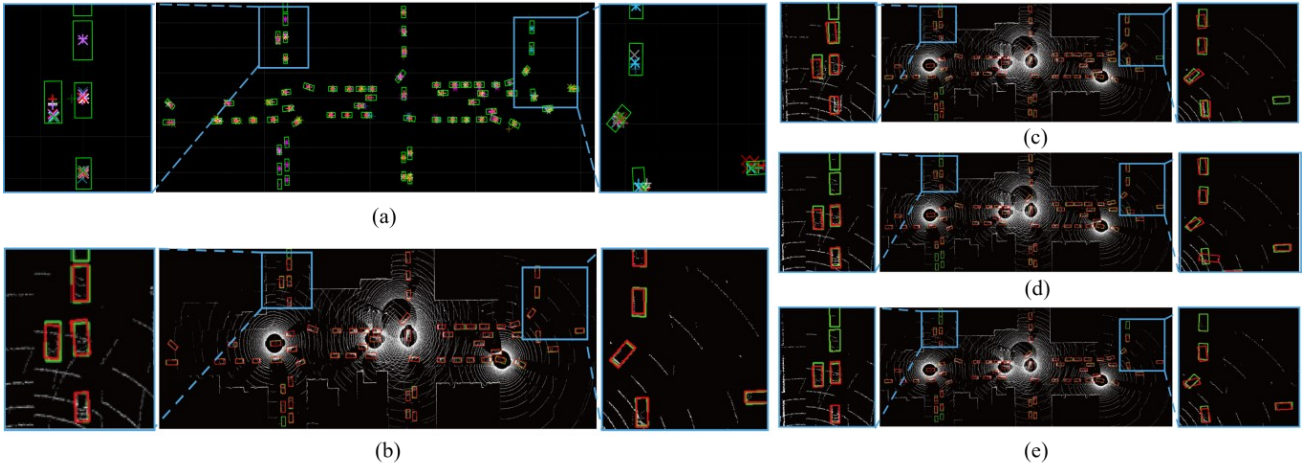

Figure 5. Visualization of (a) SemDBs and detection results of (b) Which2comm, (c) V2X-ViT, (d) Where2comm, and (e) CodeFilling on intersection scenario frame in V2XSet. Colored '+' and '×' symbols represent *SemDBs* from different agents, with '+' indicating current frame detections and '×' denoting historical frames. Red and green boxes represent the detection boxes and ground truth.

As for communication costs, *Which2comm* effectively reduces transmission volume and reaches **the order of 0.01 megabytes** (approximately AB of 13.36). Compared to the intermediate fusion methods, *Which2comm* achieves over $17/160\times$ lower communication volume than V2X-ViT (transmitting compressed features) and Where2comm (transmitting region-level sparse features). Also, due to the sparsity of SemDBs, *Which2comm* reaches the same order with CodeFilling that transmits integer indices of features approximated in the maintained codebook, while outperform on the perception performance. According to the vehicular communication standards shown in [15], only *Which2comm* and CodeFilling [5] can satisfy the current standards of 27 MB/s, (i.e., 0.338 MB, AB of 18.435 in a single timestamp). The evaluation result demonstrates that transmitting only object-level features is sufficient to achieve high-precision object detection performance.

**Latency robustness**. Asynchronous error is sampled at 100 ms intervals from 0 to 400 ms to assess models' robustness to latency, and the results are shown in Fig. 4(a). *Which2comm* demonstrates enhanced robustness to latency, particularly evident within the 0-200ms range. The enhanced performance is attributed to: 1) the temporal encoding and temporal fusion design that fuses features from the past two timestamps; 2) SemDBs that effectively integrate high-level semantic features with low-level detail representations of objects.

**Computational and parameter analysis**. We benchmarked *Which2comm* on V2XSet using a single RTX 3090 GPU, as shown in Tab. 3. *Which2comm* exhibits significant computational advantages: it employs fewer parameters and requires less memory than most competitors, owing to its sparse convolution architecture. In terms of inference speed, Which2comm achieves approximately 20Hz, outperforming V2VNet [1] and V2X-ViT [16]. Time profiling reveals that sparse convolutional blocks in backbone account for the majority of computation time, which can be further optimized through quantization and pruning for real-time applications. Detailed results can be found in Sec. 2.1 in supplementary.

### 4.3. Qualitative results

Fig. 5 visualizes the SemDBs and detection boxes projected to 2D space on V2XSet. As shown in Fig. 5(a),

Table 2. Ablation results under perfect setting and latency setting of 100ms on V2XSet. Abbreviations are utilized in the table: RTE (relative temporal encoding), SE (shuffle-ego), RW (re-weighting), RV (re-voxelization), and TF (temporal fusion).

| RTE | SE | RW | RV | TF | Perfect Setting AP@0.5 | Latency Setting AP@0.5 |
|---|---|---|---|---|---|---|
| - | - | - | - | - | 0.877 | 0.805 |
| √ | - | - | - | - | 0.868 | 0.843 |
| √ | √ | - | - | - | 0.905 | 0.891 |
| √ | √ | √ | - | - | 0.917 | 0.900 |
| √ | √ | √ | - | √ | 0.910 | 0.893 |
| √ | √ | √ | √ | - | 0.920 | 0.908 |
| √ | √ | √ | √ | √ | **0.929** | **0.919** |

where different agents are distinguished by color, SemDBs are precisely localized through our re-voxelization process. Furthermore, through temporal fusion, *Which2comm* leverages sequential information to refine detection box positions and capture object motion patterns, effectively mitigating the impact of location discrepancies on detection results. In terms of detection results shown in Fig 5(b)-(e), we observe that in areas where buildings obstruct the visibility of certain agents, perception models typically exhibit increased false negative rate, while *Which2comm* exhibits higher recall rates compared to other models. Additionally, *Which2comm* provides more precise bounding box pose estimations, particularly in right-side areas predominantly covered by a single agent's LiDAR. These advantages stem from *Which2comm's* temporal fusion mechanism, allowing it to track the motion status of objects, effectively addressing occlusion issues and enhancing the consistency of object information.

### 4.4. Ablation studies

**Ablation of individual module components**. Tab. 2 shows the ablation results conducted on V2XSet, under both perfect setting and latency setting (100 ms). The baseline model performed detection without alternative structures replacing the removed modules. Specifically, it only transmits current-frame sparse features without temporal encoding and excluded the re-weighting module, re-voxelization operation, and shuffle-ego augmentation. The results indicate that: 1) Relative temporal encoding mechanism can significantly enhance the model's performance under latency settings; 2) The shuffle-ego data augmentation strategy shows a noticeable improvement in detection performance, highlighting the importance of data augmentation in multi-agent detection tasks; 3) Sparse feature re-weighting effectively suppresses unstable features in model predictions, enhancing overall prediction accuracy; 4) Only by simultaneously adding temporal fusion and feature re-voxelization can there be a significant improvement in performance, demonstrating the effectiveness of re-voxelization in addressing aliasing effects, as well as the impact of temporal fusion on enhancing model's detection capabilities.

Table 3. Results of computational and parameter analysis.

| Method | #params (M) | Inference GPU Memory (MB) | Inference Speed (FPS) |
|---|---|---|---|
| V2VNet | 14.6 | 1996 | 15.8 |
| V2X-ViT | 14.2 | 1432 | 10.4 |
| Where2comm | 8.1 | 1701 | 48.7 |
| CoAlign | 11.1 | 2703 | 52.6 |
| **Which2comm** (p=0/p=1/**p=2**) | 10.5/10.5/**10.5** | 1296/1392/**1556** | 20.3/18.4/**19.0** |

**Ablation of temporal fusion mechanism**. Fig. 4(b) illustrates the impact of varying numbers of preceding timestamps fused during the temporal fusion process. The temporal fusion mechanism trades a slight decrease in precision under zero-latency conditions for significantly enhanced robustness to communication delays, as models incorporating historical features (*p*=1, 2) exhibit a smoother performance degradation as latency increases. Also, a higher number of historical timestamps contributes to further improved robustness. Moreover, we conducted parameter and runtime analysis on RTE mechanism. As shown in the last line of Tab. 3, incorporating historical frames introduces only marginal computational overhead. Each additional historical frame requires merely ~150MB of additional memory.

## 5. Conclusions

This paper introduces *Which2comm*, a multi-agent collaborative 3D object detection framework leveraging sparse SemDBs. By incorporating semantic information into object detection boxes, the proposed SemDB takes full advantage of late fusion and intermediate fusion approaches, allowing an object-level sparse feature transmission among agents. Moreover, we further improve model performance and latency robustness through temporal fusion methods. By designing a shuffle-ego data augmentation strategy, we enhance the utilization of multi-terminal perceptual datasets. The benchmark comparative experiments, robustness experiments, and ablation studies demonstrate that *Which2comm* achieves the SOTA performance on both perception accuracy and communication efficiency, highlighting that transmitting object-level features is sufficient to achieve high-precision and robust detection performance.

**Limitation and future work.** *Which2comm* has demonstrated its effectiveness with point clouds as input. However, it does not yet effectively utilize texture and color information from image modalities. Exploring approaches of incorporating images as input to enhance the model's perception abilities and improve detection performance across diverse object categories remains a promising research topic.